# On the Unexpected Abilities of Large Language Models

Stefano Nolfi
Institute of Cognitive Sciences and Technologies
National Research Council (CNR-ISTC)
Roma, Italy
e-mail: stefano.nolfi@cnr.it

**Abstract**
Large Language Models (LLMs) are capable of displaying a wide range of abilities that are not directly connected with the task for which they are trained: predicting the next words of human-written texts. In this article, I review recent research investigating the cognitive abilities developed by LLMs and their relation to human cognition. I discuss the nature of the indirect process that leads to the acquisition of these cognitive abilities, their relation to other indirect processes, and the implications for the acquisition of integrated abilities. Moreover, I propose the factors that enable the development of abilities that are related only very indirectly to the proximal objective of the training task. Finally, I discuss whether the full set of capabilities that LLMs could possibly develop is predictable.

## 1. Introduction

Large language models (LLMs) such as LLaMA (Touvron et al., 2023), GPT-3 (Brown et al., 2020) and PaLM (Chowdhery et al., 2022) consist of large neural networks containing hundreds of billions (or more) parameters trained on hundreds of terabytes of human-written text data. More specifically, given a sequence of tokens x = {$x_1$, ....., $x_n$}, where tokens encode words or word-parts, LLMs are trained autoregressively to predict the target token $x_i$ based on the preceding tokens $x_{<i}$. They are based on the Transformer neural network architecture (Vaswani et al., 2017) where multi-head attention layers are stacked in a very deep neural network.

After being trained on large text corpora, LLMs are endowed with several abilities. However, they do not necessarily elicit the abilities required to follow the users' instructions (Brown et al., 2020; Rae et al., 2021; Thoppilan et al., 2022). Moreover, since the training corpora include both high-quality and low-quality data, they can generate toxic, bias, and harmful content. In other words, they do not necessarily display helpful, honest, and harmless behaviors (Gehman et al., 2021; Bender, 2021; Bommassani et al., 2021; Weidinger et al., 2021; Tamkin et al., 2021). These problems can be alleviated by fine-tuning them, i.e. by continuing the training process with a relatively small set of additional training data. This data is typically consists of a list of instruction and output texts where the former denotes the human instruction to the model and the latter denotes the desired output to the instruction (Zhang et. al, 2023). The process can be realized with Instruction Finetuning and/or with Reinforcement Learning trough Human Feedback (RLHF). In the former case the training data is prepared manually and the model is trained to predict the next word autoregressively as in pre-training (Ouyang, 2022). In the latter case the training data is generated in a partially automated manner. More specifically, the system is trained to select the best option among alternative outputs produced by the system itself. The best option is identified by training a critic network to predict human preferences by using an annotated training set prepared manually (Christiano et al., 2017; Ouyang et al., 2022; Bai et al., 2022). Finetuning does not guarantee a complete solution to the problems described above but permits significant improvements in the ability to follow human instructions and the alignment of the system to human values without requiring extensive retraining or architectural changes.

The emergence of a wide range of abilities as a result of this process came as a surprise even to the developers of the LLMs themselves (Brown et al., 2020). Indeed, no one previously imagined the possibility of such a remarkable result. In this article, I illustrate some of these abilities, discuss how they are acquired, why their development was unexpected, and to what extent their appearance can be predicted. Finally, I discuss some of the differences between human and LLMs intelligence.

Alternative LLMs differ in several respects, such as the number of parameters, size and composition of the training data, training time, and procedure used to align their output to human needs and values. They differ in performance both quantitatively and qualitatively. They are released as closed or open-source software. Moreover, some LLMs, such as GPT-4 (Bubeck et al., 2023) and PaLM-E (Driess et al., 2023), process multimodal information. To discuss the general properties of these models, I will focus my analysis on the abilities that that can be induced by training these systems with text only. Where possible, I will focus the discussion in particular on the evidence collected on open-source models in which the inner workings and the details of the training process are fully transparent.

## 2. Unexpected abilities

Large Language Models (LLMs) are capable of displaying a wide range of abilities that are not directly connected with the task for which they are trained: predicting the next words of human-written texts. One of the most striking skills of LLMs is formal linguistic competence (Mahowald et al., 2023), i.e., the capacity to produce and comprehend language. These systems can produce text that is hard to distinguish from human output and are capable of correctly discriminating grammatical vs ungrammatical sentences by passing challenging tests designed by the natural language research community (Warstadt et al., 2020; Warstadt & Bowman, 2022). The acquisition of this ability is perhaps not too surprising considering that these systems are trained by using a massive collection of human-written text. On the other hand, the quality of the competence acquired largely surpasses what linguists could imagine only 5 years ago and falsifies past claims stating that statistical approaches would never be able to capture the complex syntactic and semantic features of language (Pinker & Prince, 1988; Petroni et al., 2019; Everaert et al., 2015).

LLMs trained on large text corpora also acquire large amounts of factual knowledge (Petroni et al., 2019; Robert et al., 2020; Elazar et al., 2021) which enable them to achieve state-of-the-art results on open-domain question answering benchmark without accessing external knowledge (Robert et al., 2020). Also this remarkable ability is not particularly surprising since the unstructured text used to train them contains a wealth of non-linguistic information, such as "the capital of Italy is Rome" and "two plus three is five".

In addition to formal linguistic and factual knowledge competence, LLMs display a large set of additional competences that have surprised everyone, including the developers of these systems. Below are some of the most remarkable examples.

LLMs can perform dynamical semantic operations, i.e., understanding how the meaning of a sentence alters the context described in the preceding sentences (Li et al, 2021). For example, consider the sentence "You see an open chest. The only thing in the chest is an old key. There is a locked wooden door leading east" followed by the second sentence "You pick up the key." An LLM can understand that these two sentences can be followed by the sentence "Next, you use the key to unlock the door" or "Next, you drop an apple on the ground" but cannot be followed by the sentence "Next, you remove an apple from the chest" since the previous sentences imply implicitly that the chest is empty (Li et al., 2021).

LLMs exhibit theory of mind skills that enable them to infer the mental states of the characters described in a text. Specifically, they attribute thoughts, desires, and goals to characters, posit intentions, and explain the actions of characters based on their goal (Kosinski, 2023). Although LLMs do not necessarily match human performance (Trott et al., 2023; Ulman, 2023). Still, they display remarkable abilities. For instance GPT-4 managed to solve nearly all the tasks (95%) of the 40 classic false-belief tasks that are widely used to test theory of mind skills in humans (Kosinski, 2023).

LLMs also display a certain ability to recognize affordances (Jones et al., 2022), i.e., discriminating the action that an agent can and cannot perform with an object in a given context, and to perform logical reasoning (Talmor et al., 2022; Cresswell et al., 2022), although their performance is still limited compared to humans.

LMMs learn and use representations of the outside world, at least to some extent. Indeed, they acquire internal representations of colors words that closely mirror the properties of human color perception (Abdou et al., 2021; Patel & Pavlick, 2022; Søgaard, 2023). Moreover, they can internally represent the spatial layout of the setting of a story (Patel & Pavlick, 2022; Bubeck et al., 2023) and update such representations as more related information is revealed (Li et al., 2021). They are thus capable of extracting physical knowledge indirectly from written text, without observing the world directly and without interacting with it.

LLMs can modify the text they generate in a remarkably flexible manner based on the language content provided as input. This property is referred to as in-context learning (Brown et al., 2020). It enables them to learn on the fly to perform a task described through instructions and/or examples. Although in-context learning does not produce a modification of the model's weight, it has been shown to be functionally equivalent to gradient descent (Dai et al., 2022).

For a review of known skills and performance and for an analysis of the skills that are beyond the capabilities of current language models see Srivastava et al. (2022) and Bubeck et al. (2023).

Although the development of these abilities primarily occur during the training process, the fine-tuning process plays a crucial role too. In particular, it significantly improves LLMs' ability to follow users' instructions and to produce outputs that are better aligned with human values. The analysis of the behavior shift of LLMs of the LLaMa family (Touvron et al., 2023) after fine-tuning indicate that it: (i) improves the recognition of the instruction words contained in the input prompt, generating higher-quality responses; (ii) aligns the knowledge in feed-forward layers with user-oriented tasks; and (iii) improves the detection of the interaction between word-word relation and instruction verbs (Wu et al., 2023). Regarding alignment with human values, finetuning can dramatically reduce the effect of widely recognized biases in LLMs behavior, such as anti-Black racism (Ganguli et al., 2023). Furthermore, it can be used to steer the model's behavior towards a given persona or a given set of values (Dinan et al., 2019; Bai et al., 2022a; Ganguli et al., 2022), thus demonstrating that LLMs should not necessarily express the values encoded in their training corpora (Bowman, 2023).

However, it is important to stress that fine-tuning techniques are far from perfectly effective. They cannot guarantee that the model will behave appropriately in every possible situation or that it will use all its knowledge and skill to behave as appropriately as possible. Moreover, fine-tuning techniques might have fundamental limitations arising, for example, from the fact that the probability of the existence of adversarial prompts capable of eliciting the execution of any given undesired behavior increase with prompt length (Wolf et al., 2023).

**3. Indirect acquisition of skills**

We might wonder how predicting the next words of human-written text can promote the development of a large set of complex cognitive skills. The answer is that these skills are acquired indirectly by the need to predict the next words accurately. This is because accurate prediction of the next words requires a deep comprehension of the preceding text, and this comprehension requires the possession and use of cognitive skills. Therefore, the development of the cognitive skills is induced indirectly.

Indirect acquisition processes of this kind are known to occur in other adaptive processes, such as natural evolution and individual learning. The body structure of bacteria, plants, animals, and humans evolved as a side effect of the attempt of small molecules to replicate as efficiently as possible. Moreover, the behavioral skills developed by natural organisms, such as walking, flying, escaping predators, communicating, etc., also evolved as a side effect of the attempt of these molecules to replicate as efficiently as possible. Thus also in the case of natural evolution, structural properties and skills emerge indirectly as a result of attempt to maximize a different ability. The spontaneous acquisition of skills that are not rewarded directly but that are instrumental to a function that is rewarded directly is observed also in artificial evolving systems. For example, the ability to communicate through a self-organizing communication system emerges spontaneously in groups of robots selected for the ability to forage (Nolfi & Mirolli, 2010). Finally, indirect acquisition processes characterize individual learning too. For example, dexterous manipulation skills that enable a robotic hand to rotate and translate objects emerge spontaneously in robots trained to solve the Rubik's cube problem (Akkaya et al., 2019). Clearly, these indirect processes differ in several important ways. For instance, natural evolution operates at the level of populations of individuals and at multiple organizational levels, while individual learning does not. Additionally, natural evolution operates on a much larger timescale than individual learning. Furthermore, populations of individuals adapt thanks to the variations occurring during reproduction and to the selection of fittest individuals, which is a process that is qualitatively different from the process through which LLMs learn. However, they all demonstrate the possibility to generate a wide range of capabilities as a side effect of the attempt to maximize a given capability or skill.

Another indirect acquisition process worth mentioning is the acquisition of word meanings in humans (Mollo & Millière, 2023). In many cases the meaning of words is not acquired through direct experience but through representations produced by others who may have learned them in the same way without necessarily interacting directly with the corresponding world referents. "For example, most people have never been in direct casual contact with enriched uranium, but that does not entail that their through and statements about uranium lack referential grounding---they are indeed about uranium." (Mollo & Millière, 2023, p.21).

As an example of indirect skill acquisition in LLMs, let's consider theory of mind and reasoning skills. To predict as accurately as possible the words that the characters of a story will say next, the system should infer the goals of the characters from their behavior and should differentiate between the events that happened in the story and the events of which the characters are aware of. In other words, the system should acquire an ability to reason and to identify the mental states of the characters and an ability to imagine the actions that they can take to achieve their own goals.

Clearly, the more indirect the relationship is between the skills that need to be acquired and the task that is rewarded directly, the lower the probability that the skills are acquired. From this point of view, the development of complex cognitive skills that are related very indirectly to the task of predicting the next words still results surprising. The successful acquisition of those skills can be explained by considering a set of enabling factors that characterize the LLMs domain.

A first enabling factor is the high informative nature of the prediction error, i.e., the fact that it provides a very reliable measure of the knowledge and skills of the system. This implies that improvements and regressions of the system skills always lead to decreases and increases of the error, respectively, and vice versa. A second enabling factor is the predictability of human language granted by its symbolic and non-dynamical nature. The predictability of dynamical systems, such as robots interacting physically with their external environment, is limited by the complex system nature of their dynamics. This is due to the fact that minor differences tend to produce large effects over time. The task of predicting the next word benefits from the absence of this limitation that is granted by the non-dynamical nature of language. Please note that I am not claiming that language is non-dynamical as a whole. I am referring here only to the spoken or written expression of language, i.e., to the aspects of language that are relevant for the task of guessing the next words. A third enabling factor is the availability of a huge set of data ready to be used for training. Indeed, the remarkable ability of LLMs manifests only when their size and training set are scaled-up to huge dimensions (Kaplan et al., 2020). Overall, these factors can explain why LLMs manage to acquire cognitive skills that are related in a very indirect manner to the task of predicting the next words.

Indirect acquisition of cognitive skills also allows for achieving another remarkable result: the acquisition of integrated capabilities, i.e., skills that are organized to work in synergy with other acquired skills. This is because the utility of each cognitive skill for the prediction task depends on the extent to which such skill is integrated with other existing skills. In other words, skills acquired indirectly become naturally integrated with other indirectly acquired skills. As an example, we can consider the tight integration of the factual knowledge and the linguistic competence of LLMs. The integrated nature of these capabilities allow to recover the knowledge possessed by LLMs through natural language queries (Roberts et al. 2020). Another example is the acquisition of integrated theory of mind and reasoning skills which enable to infer the goals of the characters of a story from their behavior and to guess the actions that the characters might do to achieve their goals (Kosinski, 2023).

Embodied theories of language acquisition in humans (Cisek, 1999; Kolodny & Edelman, 2018) postulate a similar indirect acquisition process. More specifically, they postulate that language originates in humans as a form of action that has the function of influencing the behaviors of the conspecifics (and perhaps also of the self [Mirolli & Parisi, 2019]). It starts with the production of a crying behavior that allow to obtain the parent's attention and care and proceeds with the discovery of alternative vocalizations that enable to obtain alternative specific intended outcomes. Interestingly, this implies that although the acquisition of language and associated skills in humans and LLMs differ in important respects (which I will discuss in Section 5) it might rely on an indirect process in both cases.

**4. Predictability and emergence**

An important issue that needs to be investigated is how well the skills acquired by LLMs are predictable. This question has important implications for AI safety and alignment, since the impossibility of predicting the abilities that will be developed by larger models implies that these models could acquire undesired and dangerous capabilities, without warning (Schaeffer et al., 2023).

The performance of large language models scales as a power-law with model size, dataset size, and amount of computation used for training (Kaplan et al., 2020). This implies that the overall performance (prediction error) of these systems is predictable. In other words, it implies that the overall performance that can be obtained by increasing the size of the model and/or the training time can be extrapolated based on the performance displayed by models that are smaller or less trained. However, the specific abilities that will be developed by a model of a certain size are not necessarily predictable.

The interest in this topic was raised by the publication of an influential article by Wei et al. (2022) entitled "Emergent abilities of large language models". The authors observe that several abilities, such as in-context learning, instruction following, step-by-step reasoning, and arithmetic skills, are not present in smaller models. They appear in larger models only. Moreover, they show that the acquisition process of these abilities in sufficiently large models is characterized by a sharp transition between a phase in which the system does not show any progress in the ability and a subsequent phase in which the system progresses. According to the authors, these two observations imply that predicting the abilities that will be acquired by scaled-up models will be impossible. In other words, they claim that the abilities that will be developed by scaled-up models cannot be extrapolated from what we know about smaller models.

Wei et al. (2022) later questioned the presence of sharp transitions during the training process and showed that the occurrence of sharp or more continuous transitions in the acquisition process depends on the metric used to evaluate the performance. The occurrence of sharp transitions might thus be induced by the utilization of specific metrics.

Here, I would like to distinguish two aspects of predictability: (i) predicting the specific abilities that an LLM of a certain size trained on a corpus of a given size will develop, and (ii) predicting the full set of capabilities that LLMs could possibly develop. In principle, the abilities that could be acquired by LLMs of a sufficiently large size could be unbounded. Indeed, the possibility of generating an open-ended process that keeps producing newer and newer capacities without ever reaching a bound characterizes natural evolution (Taylor, 2019). However, the characteristics of the process through which LLMs acquire their skills suggest that the list of skills they can acquire is bounded and is restricted to the set of abilities possessed by the humans who wrote the text used to train the models. If this hypothesis holds, we should expect that models trained to predict the human written text would not develop alien abilities, i.e., abilities unknown to humankind. The reason why the abilities necessary to comprehend human-written text are limited to the abilities possessed by humans is that human language is an artifact of humans themselves that has been shaped in its form by the cognitive abilities of human speakers. The same reason could explain why the cognitive abilities possessed by humans should be sufficient to comprehend human language. However, notice that the hypothesis that the abilities that could be developed are bounded by the abilities possessed by humans does not resolve the safety issue mentioned above. Indeed, this does not rule out the possibility of achieving super-human performance, as discussed in the next section, and does not rule out the possibility of exhibiting undesiderable behaviors.

This contrasts with other machine learning methods and domains in which the set of solutions that could be found is potentially unbounded and is not constrained by the solutions possessed by humans. In particular, it contrasts with trial-and-error learning methods in which the learning system attempts to maximize a utility function by interacting actively with an external environment. For example, it contrasts with AlphaZero (Silver et al., 2017), a system that learns to play the game of GO by playing against itself. The training is realized by introducing random variations in the actions produced by the system and by increasing the probability to produce the varied actions that increase the chances of winning the game. Indeed, the analysis of the behavior discovered by this system revealed strategies that were previously unknown to humans (Silver et al., 2017). Notice that, by contrast, LLMs do not interact with an external environment. They are passively exposed to a sequence of words, and they do not have the possibility to alter their next observations through their actions. Moreover, they do not learn by trial-and-error. They learn by minimizing the offset between the words predicted and the words that actually follow.

Future studies investigating the course of the learning process in LLMs can improve our understanding of the dependency relationships among abilities and shed light on the order in which they tend to be developed. The preliminary analysis reported in Srivastava et al. (2022) indicates that skills having a compositional nature progress only after the acquisition of the required sub-skills. For example, LLMs acquire an ability to identify chess moves corresponding to mates only after they develops a good ability to discriminate between valid and invalid moves (Srivastava et al., 2022). Another type of study that could provide insights into the developmental course concerns comparing the outcome of multiple training sessions of the same model.

**5. LLMs versus Natural Intelligence**

LMMs differ from humans in several respects.

Probably the most striking difference is that humans acquire much of their knowledge and skills actively by interacting sub-symbolically with the external environment and by interacting sub-symbolically and symbolically with other humans. LLMs instead acquire much of their knowledge and skills passively by being exposed to symbolic information only (human written text). This difference triggered a heated debate between those who think that LLMs use words without really understanding their meaning and those who think that they have a genuine comprehension (Mitchell & Krakauer, 2023). The former group stresses the limitations of current LLMs, e.g. their difficulty with causal and multi-step compositional reasoning (Dziri et al., 2023), their limited sensitivity to affordances with respect to humans (Jones et al., 2022), and their inability to distinguish their own output from factual knowledge (Ortega et al., 2021). More generally, they claim that they could not have a real understanding since they do not have any direct experience of the world they talk about. The latter group, on the other hand, stresses that: current models display already remarkable performance and that current limitations are likely to be overcome in future models. More generally, they claim that the knowledge of the physical world can be extracted indirectly from the traces left in human written text. The possibility to extract knowledge indirectly is supported by evidence collected on colorblind people who associate colors to the same emotions as sighted people (Saysani et al., 2021). Moreover, the fact that language contains sufficient information is supported by the fact that LLMs trained also with images do not acquire better representations than models trained with language data only (Yun et al., 2021). Whether or not acquiring knowledge passively, without interacting with the external environment, and acquiring knowledge from language input only limits the quality of the representations that can be acquired thus represents an open question for the moment (Pezzulo et al., 2023). Similarly, the relative importance that direct and indirect knowledge acquisition have in the case of humans still represents an open question (Borghi et al., 2023).

A second difference concerns the amount of training data. State-of-the-art LLMs are exposed to language corpora that are much larger and wider in content than those experienced by a typical human (Warstadt et al., 2023). By the age of 10, when children acquire a mature linguistic competences, they have been exposed to a total of 30-110 million words, while LLMs such as GPT-3 are exposed to more than 200 billion words. Moreover, LLMs are trained with text covering all known scientific disciplines. The impact of these differences is hard to anticipate. The use of huge training corpora might be necessary to compensate the inability of LLMs to interact with the physical and social environment directly and/or the lower quality of the data. Consequently, it might not necessarily lead to super-human performance. The use of training data covering a wide range of topics and knowledge could enable LLMs to exceed the skills possessed by any single human. Moreover, the acquisition of a vast set of knowledge and skills could amplify the beneficial effects that can be gained through transfer learning. Indeed, the possibility to transfer knowledge among skills could permit surpassing the quality of the training examples. An example of transfer learning of this kind is reported in the experiments reported by Lee et al. (2022) in which the authors trained a single transformer network to play N different Atari games by imitating the strategy of N players trained on N different games through reinforcement learning. The system trained to perform multiple games managed to outperform some of its teachers on some corresponding games.

A third difference concerns computational properties. The human brain has a much greater number of neurons and connections than state-of-the-art LLMs. Moreover, natural neurons are much more complex than artificial neurons. On the other hand, the transformer architecture used by LLMs can process thousands of words at once without any loss of information, while human brains process sequences of items sequentially and have severe limitations in short-term memory. Humans thus face the now-and-never bottleneck (Christiansen & Carter, 2016) that forces them to process linguistic input as rapidly as possible, while LLMs do not. Clearly, these differences can have a significant impact on performance and the way in which linguistic information is processes. For an analysis of the consequences of the bottleneck in humans, see (Christiansen & Carter, 2016).

A fourth difference concerns the fact that individual humans have specific values, beliefs, goals, and desires while LLMs do not. Contrary to what could be expected, however, LLMs do not necessarily express the values encoded in the text used to train them, especially when the behavior of the pretrained model is steered through prompting or fine-tuning (Bowman, 2023). In other words, the values expressed by a LLM do not necessarily reflect the average of the values expressed in its training data. Indeed, exposing models to more examples of unwanted behavior during pretraining can improve their ability to avoid the production of such behavior through fine-tuning (Korbak et al., 2023). Due to the passive nature of their training process, LLMs cannot acquire their own values and goals autonomously. However, they can be prompted or fine-tuned to exhibit the behavior of a specific chosen persona (Chuang et al., 2023), although not necessarily in full (Cheng et al., 2023). Moreover, if allowed to interact with other LLMs, they are capable of modifying their initial belief (Chang et al., 2023).

Finally, detailed comparisons that take into account not only the performance level but also the way in which problems are solved are revealing qualitative differences between LLMs and humans. For instance, GPT-3 matches human behavior in decision from description problems in which the outcome probability of the alternative options is fully described. However, it displays a qualitatively different behavior in decision from experience problems in which the decision-maker has to infer the probability of the alternative outcomes from multiple observations (Binz & Schulz, 2023a, 2023b).

## 6. Conclusion

The emergence of a series of cognitive abilities in transformer neural networks trained to predict the next words of human-written text came as a surprise even to the developers of LLMs. Nobody anticipated such a remarkable result, namely that the development of several linguistic and cognitive abilities could be induced indirectly by attempting to guess the next words as accurately as possible.

We were aware of indirect acquisition processes of this kind. We know that in natural evolution, the attempt to maximize the probability of reproducing led to the development of a wide range of species possessing many skills. Moreover, we know that in trial-and-error learning processes, the acquisition of a given skill can induce the development of abilities that are instrumental to that skill. On the other hand, nobody hypothesized that a similar indirect process could be obtained by attempting to predict the next words of written text. As claimed in this article, this surprising result can be explained by considering the non-dynamical nature of language data and the highly informative nature of the prediction error. The non-dynamical nature of language rules out the problem affecting dynamical systems, namely the fact that minor variations at a certain state can have huge effects over time. The highly informative nature of the prediction error ensures the efficacy of the learning process and also enables the acquisition of skills that are related in a very indirect manner to the prediction task. More precisely, these enabling factors permit to reconstruct the cognitive abilities possessed by the humans who wrote the text corpus from the text corpus itself.

As claimed in Section 3, the indirect nature of the acquisition process also allows for the development of well-integrated solutions, i.e., solutions in which each acquired ability is shaped to work in synergy with the other acquired skills. The exploitation of an indirect acquisition process thus plays a crucial role. Interestingly, an indirect acquisition process seems to characterize human development too. Indeed, humans acquire communication and cognitive skills, at least in part, in an attempt to alter the behaviors of their conspecifics with the purpose of achieving their own goals. Although the development of cognitive abilities in humans and LLMs differs in several important respects, as discussed above, it might share this indirect property.

Finally, I discussed the extent to which the abilities that can be developed by LLMs are predictable. I hypothesized that the indirect acquisition of these abilities implies that the set of abilities that can be acquired is restricted to the set that is necessary and sufficient to understand human-written text. Consequently, the set of abilities that can be acquired is probably restricted to the set of abilities possessed by the humans who wrote the text used for training. This hypothesis does not contradict the possibility of achieving super-human performance. Indeed, the integration of knowledge and skills possessed by multiple humans that exceed those possessed by any single human and the ability to process long sequences of items without information loss can allow LLMs to outperform the abilities of individual humans.

## Acknowledgment

I acknowledge financial support from PNRR MUR project PE0000013-FAIR.